\pdfoutput=1
\documentclass[11pt]{article}

\usepackage[]{EMNLP2023}

\usepackage{times}
\usepackage{latexsym}

\usepackage[T1]{fontenc}
\usepackage[utf8]{inputenc}

\usepackage{microtype}

\usepackage{tabularx}
\usepackage{tabu}
\usepackage{booktabs}
\usepackage{multirow}

\usepackage{graphicx} 
\usepackage{float}
\usepackage{subfigure} 
\usepackage{amssymb}

\usepackage{hyperref}
\usepackage{tablefootnote}
\usepackage{xspace}

\usepackage{float}
\usepackage{amsmath}
\usepackage{bm}
\usepackage{algorithmicx}
\usepackage{algorithm}
\usepackage{algpseudocode}

\usepackage{arydshln}

\usepackage{amssymb}
\usepackage{pifont}
\newcommand{\cmark}{\ding{51}}%
\newcommand{\xmark}{\ding{55}}%
\usepackage{enumitem}

\newcommand{\loft}{\textsc{LoFT}\xspace}
\newcommand{\logicnlg}{\textsc{LogicNLG}\xspace}
\newcommand{\lotnlg}{\textsc{LoTNLG}\xspace}
\newcommand{\tulu}{\textsc{T\"ulu}\xspace}
\newcommand{\wtqnew}{F2WTQ\xspace}

\newcommand{\rqone}{How do LLMs perform in table-to-text generation tasks?}

\newcommand{\rqtwo}{Can we use LLMs to assess factual consistency of table-to-text generation?}

\newcommand{\rqthree}{How can fine-tuned models benefit from LLMs' strong table-to-text abilities?}

\newcommand{\down}[1]{\textcolor{red}{#1}}
\newcommand{\up}[1]{\textcolor{green!55!black}{#1}}

\title{Investigating Table-to-Text Generation Capabilities of LLMs in\\Real-World Information Seeking Scenarios}

\author{Yilun Zhao\thanks{~~Equal Contributions.}~~$^{1}$ \quad Haowei Zhang$^{*~2}$ \quad Shengyun Si$^{*~2}$ \\ \bf{Linyong Nan$^1$ \quad Xiangru Tang$^1$ \quad Arman Cohan$^{1,3}$} \vspace{4pt}\\
$^1$Yale University, $^2$Technical University of Munich, $^3$Allen Institute for AI\vspace{4pt}\\
\texttt{yilun.zhao@yale.edu} \quad \texttt{\{haowei.zhang, shengyun.si\}@tum.de}
}

\begin{document}
\maketitle
\begin{abstract}
Tabular data is prevalent across various industries, necessitating significant time and effort for users to understand and manipulate for their information-seeking purposes. 
The advancements in large language models (LLMs) have shown enormous potential to improve user efficiency. However, the adoption of LLMs in real-world applications for table information seeking remains underexplored.
In this paper, we investigate the table-to-text capabilities of different LLMs using four datasets within two real-world information seeking scenarios. These include the \logicnlg and our newly-constructed \lotnlg datasets for \emph{data insight generation}, along with the FeTaQA and our newly-constructed \wtqnew datasets for \emph{query-based generation}. 
We structure our investigation around three research questions, evaluating the performance of LLMs in table-to-text generation, automated evaluation, and feedback generation, respectively. 
Experimental results indicate that the current high-performing LLM, specifically GPT-4, can effectively serve as a table-to-text generator, evaluator, and feedback generator, facilitating users' information seeking purposes in real-world scenarios. 
However, a significant performance gap still exists between other open-sourced LLMs (e.g., \tulu and LLaMA-2) and GPT-4 models. Our data and code are publicly available at \url{https://github.com/yale-nlp/LLM-T2T}.
\end{abstract}
\begin{figure}[!t]
    \centering
    \includegraphics[width = \linewidth]{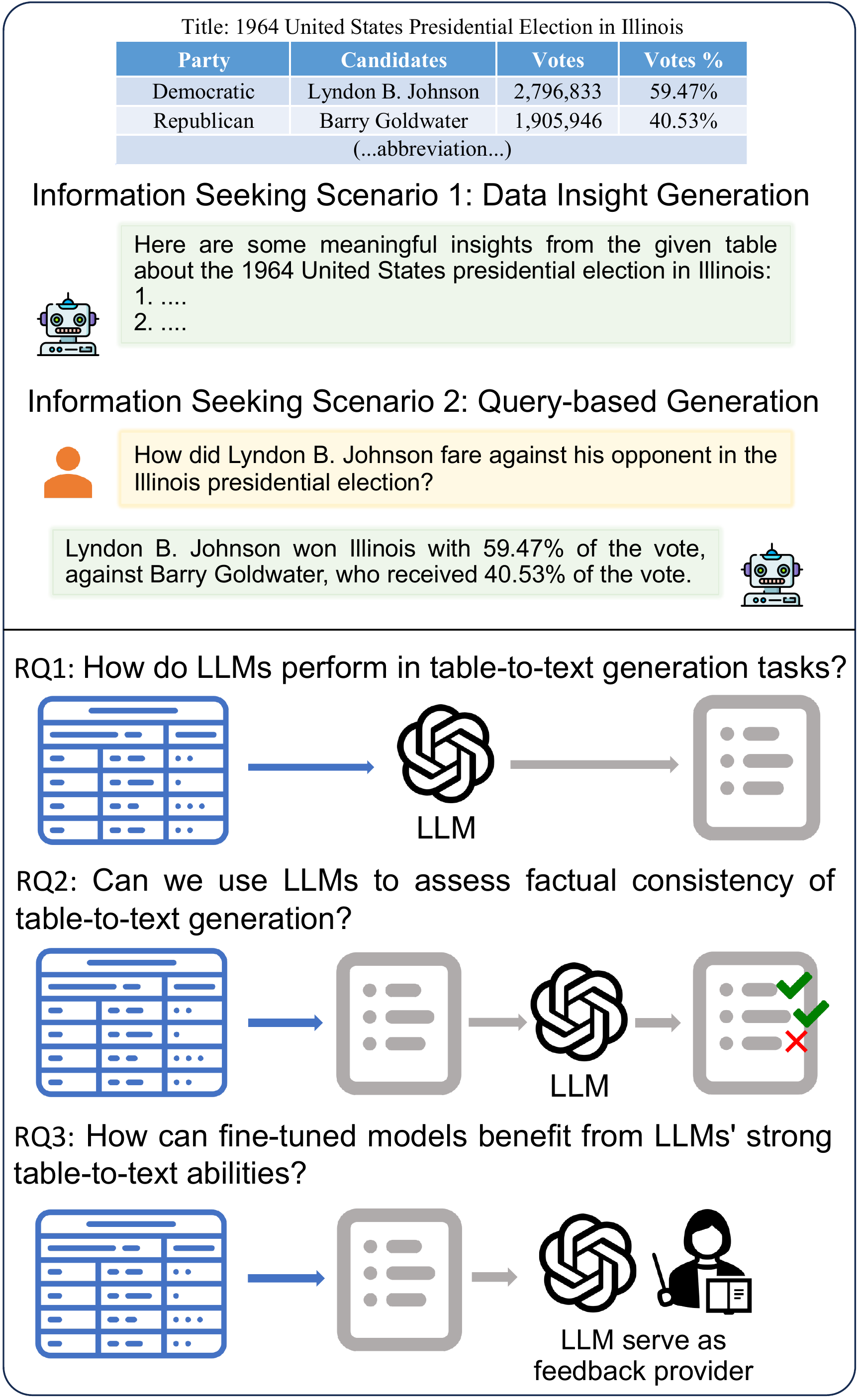}
    \caption{The real-world table information seeking scenarios and research questions investigated in this paper.}
    \vspace{-15pt}
    \label{fig:example}
\end{figure}
\begin{table*}[!t]
    \centering
    \small
    \begin{tabular}{lrrlc}
        \toprule
        \textbf{Dataset} & \textbf{\# Table} & \textbf{\# Examples} & \textbf{Control Signal} & \textbf{Rich in Reasoning?} \\
        \midrule 
        \multicolumn{5}{c}{\emph{Data Insight Generation}} \\ \\
        \logicnlg~\cite{chen-etal-2020-logical} & 862 & 4,305 & None & \cmark\\
        \textbf{\lotnlg} (ours) & 862 & 4,305 & Reasoning type & \cmark\\
        \midrule

        \multicolumn{5}{c}{\emph{Query-based Generation}} \\ \\
        FeTaQA~\cite{parikh-etal-2020-totto} & 2,003 & 2,003  & User query & \xmark\\
        \textbf{F2WTQ} (ours) & 4,344 & 4,344 & User query  & \cmark\\
        
        \bottomrule
    \end{tabular}
    
    \caption{Experimental dataset statistics for the test set. Examples of our newly-constructed \lotnlg and \wtqnew datasets are displayed in Figure~\ref{fig:lotnlg_example} and \ref{fig:wtqnew_example}, respectively.}
    \label{tab:dataset}
\end{table*}
\section{Introduction}
In an era where users interact with vast amounts of structured data every day for decision-making and information-seeking purposes, the need for intuitive, user-friendly interpretations has become paramount~\cite{Zhang2023DataCopilotBB, zha2023tablegpt, Li2023SheetCopilotBS}. Given this emerging necessity, table-to-text generation techniques, which transform complex tabular data into comprehensible narratives tailored to users' information needs, have drawn considerable attention~\cite{parikh-etal-2020-totto, chen-etal-2020-logical, nan-etal-2022-fetaqa, zhao2023qtsumm}. These techniques can be incorporated into a broad range of applications, including but not limited to game strategy development, financial analysis, and human resources management. However, existing fine-tuned table-to-text generation models~\cite{nan-etal-2022-r2d2, liu2022tapex, liu-etal-2022-plog, zhao-etal-2023-loft} are typically task-specific, limiting their adaptability to real-world applications.

The emergence and remarkable achievements of LLMs~\cite{gpt-3, scao2022bloom, wang2023tulu,Scheurer2023TrainingLM, OpenAI2023GPT4TR, touvron2023llama, alpaca, Touvron2023Llama2O} have sparked a significant transformation in the field of controllable text generation and data interpretations~\cite{nan-etal-2021-dart, zhang2022macsum, goyal2022news, koksal2023longform, gao2023human, madaan2023self, zhou2023context}. 
As for table-based tasks, recent work~\cite{chen-2023-large, Ye2023LargeLM, Gemmell2023GenerateTA} reveals that LLMs are capable of achieving competitive performance with state-of-the-art fine-tuned models on table question answering~\cite{pasupat-liang-2015-compositional, nan-etal-2022-fetaqa} and table fact checking~\cite{2019TabFactA, gupta-etal-2020-infotabs}. However, the potential of LLMs in generating text from tabular data for users' information-seeking purposes remains largely underexplored.

In this paper, we investigate the table-to-text generation capabilities of LLMs in two real-world table information seeking scenarios: 1) \textbf{Data Insight Generation}~\cite{chen-etal-2020-logical}, where users aim to promptly derive significant facts from the table, anticipating the systems to offer several data insights; and 2) \textbf{Query-based Generation}~\cite{pasupat-liang-2015-compositional, nan-etal-2022-fetaqa}, where users consult tables to answer specific questions. 
To facilitate a rigorous evaluation of LLM performance, we also construct two new benchmarks: \textbf{\lotnlg} for data insight generation conditioned with specific logical reasoning types; and \textbf{\wtqnew} for free-form question answering that requires models to perform human-like reasoning over Wikipedia tables.

We provide an overview of table information seeking scenarios and our main research questions in Figure~\ref{fig:example}, and enumerate our findings as follows:

\begin{itemize} [leftmargin=*]
\item[] \textbf{RQ1}: \emph{\rqone} \\
\textbf{Finding}: LLMs exhibit significant potential in generating coherent and faithful natural language statements based on the given table. For example, GPT-4 outperforms state-of-the-art fine-tuned models in terms of faithfulness during both automated and human evaluations. The statements generated by GPT-3.5 and GPT-4 are also preferred by human evaluators. However, a significant performance gap still exists between other open-sourced LLMs (e.g., Vicuna and LLaMA-2) and GPT-* models, especially on our newly-constructed \lotnlg and \wtqnew datasets.

\item[] \textbf{RQ2}: \emph{\rqtwo} \\
\textbf{Finding}: LLMs using chain-of-thought prompting can serve as reference-free metrics for table-to-text generation evaluation. These metrics demonstrate better alignment with human evaluation in terms of both fluency and faithfulness.

\item[] \textbf{RQ3}: \emph{\rqthree} \\
\textbf{Finding}: LLMs that utilize chain-of-thought prompting can provide high-quality natural language feedback in terms of factuality, which includes explanations, corrective instructions, and edited statements for the output of other models. The edited statements are more factually consistent with the table compared to the initial ones.
\end{itemize}
\section{Table Information Seeking Scenarios}
Table~\ref{tab:dataset} illustrates the data statistics for the four datasets used in the experiments. We investigate the performance of the LLM in the following two real-world table information-seeking scenarios.

\subsection{Data Insight Generation}
Data insight generation is an essential task that involves generating meaningful and relevant insights from tables. By interpreting and explaining tabular data in natural language, LLMs can play a crucial role in assisting users with information seeking and decision making. This frees users from the need to manually comb through vast amounts of data. We use the following two datasets for evaluation.

\subsubsection{\logicnlg Dataset} 
The task of \logicnlg~\cite{chen-etal-2020-logical} involves generating five logically consistent sentences from a given table. It aims to uncover intriguing facts from the table by applying various logical reasoning operations (e.g., count and comparison) across different table regions.

\subsubsection{\lotnlg Dataset} 
Our preliminary experiments revealed that when applied to the \logicnlg dataset, table-to-text generation systems tend to generate multiple sentences that employ the same logical reasoning operations. For instance, in a 0-shot setting, the GPT-3.5 model is more inclined to generate sentences involving numerical comparisons, while overlooking other compelling facts within tables. This lack of diversity in data insight generation poses a significant limitation because, in real-world information-seeking scenarios, users typically expect systems to offer a variety of perspectives on the tabular data. To address this issue, application developers could tailor the table-to-text generation systems to generate multiple insights that encompass different logical reasoning operations~\cite{Perlitz2022DiversityET, zhao-etal-2023-loft}.
In order to foster a more rigorous evaluation of LLMs' abilities to utilize a broader range of logical reasoning operations while generating insights from tables, we have developed a new dataset, \lotnlg, for logical reasoning type-conditioned table-to-text generation. In this setup, the model is tasked with generating a statement by performing the logical reasoning operations of the specified types on the tables. 

\begin{figure}[!t]
    \centering
    \includegraphics[width = \linewidth]{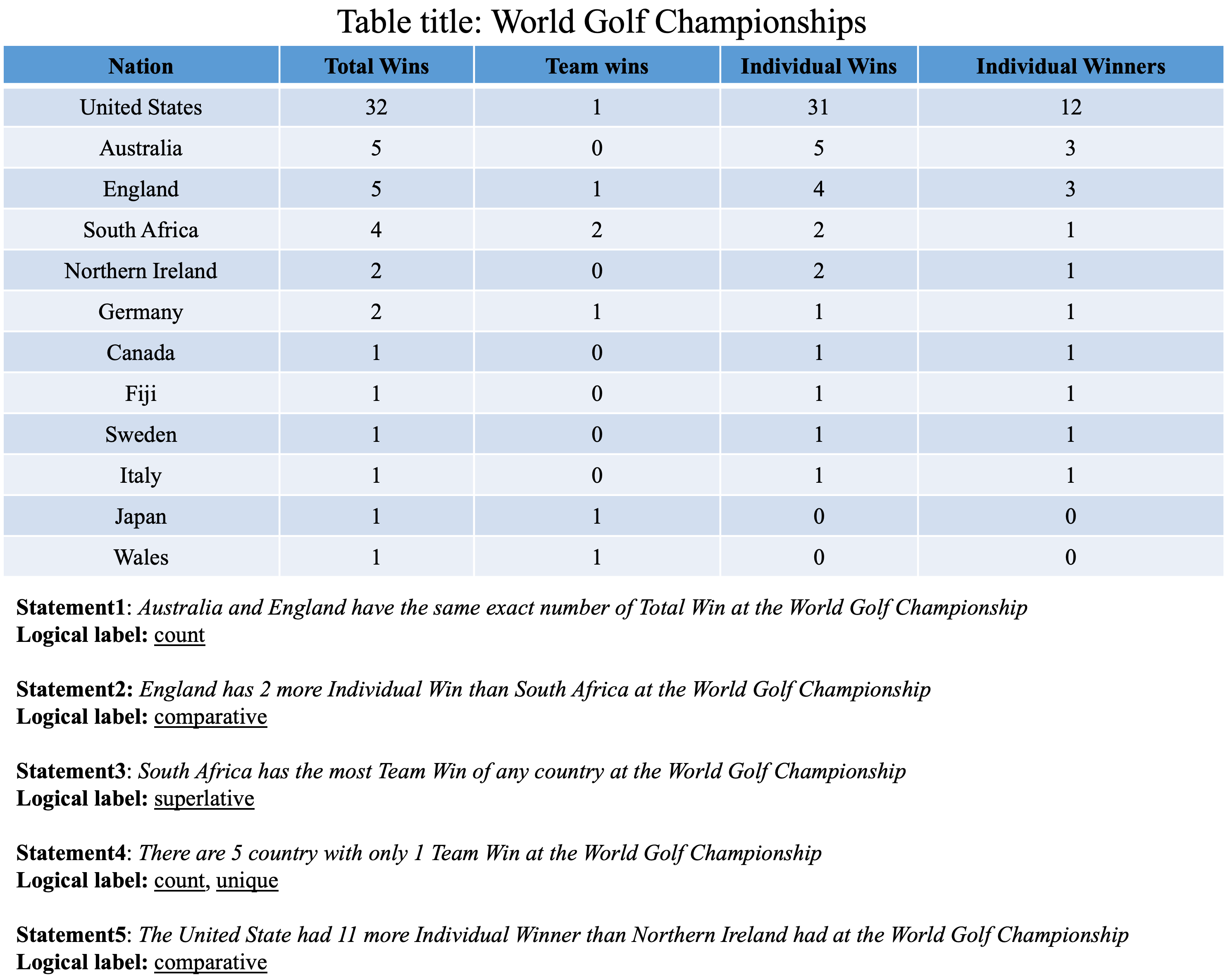}
    \caption{An example of \lotnlg, where models are required to generate statements using the specified types of logical reasoning operations}
    \label{fig:lotnlg_example}
\end{figure}

\paragraph{\lotnlg Dataset Construction}
Following \citet{2019TabFactA}, we have predefined nine types of common logical reasoning operations (e.g., count, comparative, and superlative), with detailed definitions provided in Appendix~\ref{appendix:type}. We use examples from the \logicnlg test set to construct \lotnlg. Specifically, for each statement from \logicnlg, we assign two annotators to independently label the set of logical reasoning types used in that statement, ensuring that no more than two types were identified per statement. If there are discrepancies in the labels, an expert annotator is brought in to make the final decision. The distribution of logical reasoning types in \lotnlg is illustrated in Figure~\ref{fig:type_distribution} in Appendix~\ref{appendix:type}.

\subsection{Query-based Generation}
Query-based table-to-text generation pertains to producing detailed responses based on specific user queries in the context of a given table.
The ability to answer users' queries accurately, coherently, and in a context-appropriate manner is crucial for LLMs in many real-world applications, such as customer data support and personal digital assistants. We utilize following two datasets to evaluate LLMs' efficiency in interacting with users and their proficiency in table understanding and reasoning.

\subsubsection{FeTaQA Dataset} \citet{nan-etal-2022-fetaqa} introduces a task of free-form table question answering. This task involves retrieving and aggregating information from Wikipedia tables, followed by generating coherent sentences based on the aggregated contents. 

\begin{figure}[!t]
    \centering
    \includegraphics[width = \linewidth]{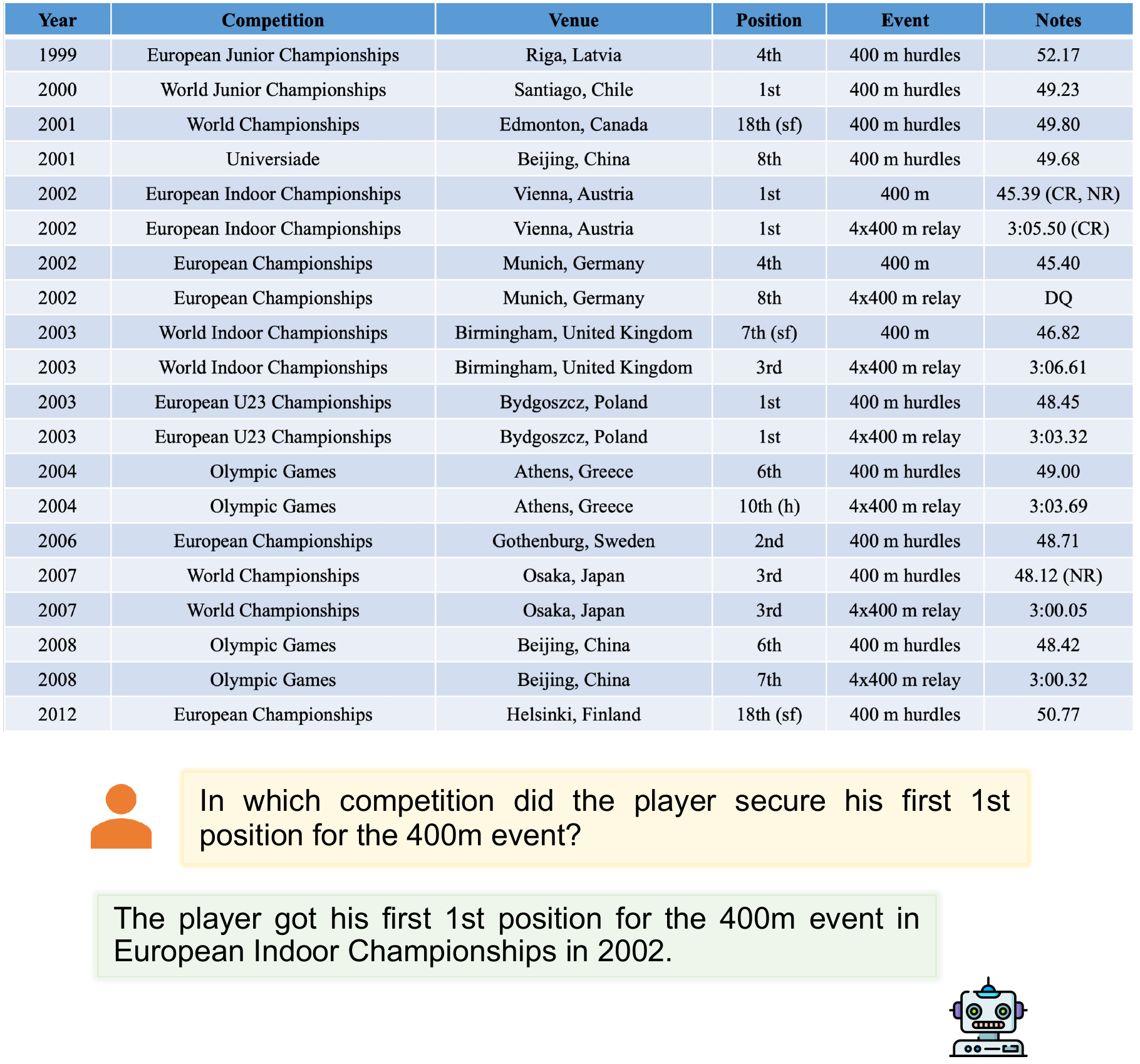}
    \caption{An example of \wtqnew, where models need to perform human-like reasoning to generate response.}
    \label{fig:wtqnew_example}
\end{figure}

\subsubsection{\wtqnew Dataset}
Queries in the FeTaQA dataset typically focus on \emph{surface-level facts} (e.g., "Which country hosted the 2014 FIFA World Cup?"). However, in real-world information-seeking scenarios, users are likely to consult tables for more complex questions, which require models to perform human-like reasoning over tabular data. Therefore, we have constructed a new benchmark, named \wtqnew, for more challenging, free-form table question answering tasks.

\paragraph{\wtqnew Dataset Construction} 
We adopt the WTQ dataset~\cite{pasupat-liang-2015-compositional} as a basis to construct \wtqnew. The WTQ dataset is a short-form table question answering dataset, which includes human-annotated questions based on Wikipedia tables and requires complex reasoning. However, we do not directly use WTQ for LLM evaluation because, in real-world scenarios, users typically prefer a natural language response over a few words. In the development of \wtqnew, for each QA pair in the WTQ test set, we assign an annotator who assumes the role of an agent that analyzes the table and provides an expanded, sentence-long response. We found that the original questions in the WTQ dataset occasionally contained grammatical errors or lacked a natural linguistic flow. In these cases, the annotators are required to rewrite the question to ensure it was fluent and natural.
\section{Evaluation System}\label{sec:eval_def}
\subsection{Automated Evaluation}\label{sec:auto_eval_definition}
We adopt following popular evaluation metrics for automated evaluation:
\begin{itemize} [leftmargin=*]
\itemsep0em 
\item\textbf{BLEU}~\cite{papineni-etal-2002-bleu} uses a precision-based approach, measuring the n-gram matches between the generated and reference statements.

\item\textbf{ROUGE}~\cite{lin2004rouge} uses a recall-based approach, and measures the percentage of overlapping words and phrases between the generated output and reference one.

\item\textbf{SP-Acc}~\cite{chen-etal-2020-logical} extracts the meaning representation from the generated sentence and executes it against the table to verify correctness.

\item\textbf{NLI-Acc}~\cite{chen-etal-2020-logical} uses TableBERT fine-tuned on the TabFact dataset~\cite{2019TabFactA} as faithfulness classifier.

\item\textbf{TAPAS-Acc}~\cite{liu-etal-2022-plog} uses TAPAS~\cite{herzig-etal-2020-tapas} fine-tuned on the TabFact dataset as the backbone.

\item\textbf{TAPEX-Acc}~\cite{liu-etal-2022-plog} employs TAPEX~\cite{liu2022tapex} fine-tuned on the TabFact dataset as the backbone. Recent works~\cite{liu-etal-2022-plog, zhao-etal-2023-loft} have revealed that NLI-Acc and TAPAS-Acc is overly positive about the predictions, while TAPEX-Acc serves as a more reliable faithfulness-level metric.

\item\textbf{Exact Match \& F-Score for Logical Reasoning Type}\quad For \lotnlg evaluation, the exact match measures the percentage of samples with all the labels classified correctly, while the F-Score provides a balanced metric that considers both type I and type II errors.

\item\textbf{Answer Accuracy} refers to the proportion of correct predictions out of the total number of predictions in \wtqnew generation. 
\end{itemize}

\subsection{Human Evaluation} \label{sec:human_eval_definition}
To gain a more comprehensive understanding of the system's performance, we also conduct human evaluation. Specifically, the generated statements from different models are evaluated by humans based on two criteria: \emph{faithfulness} and \emph{fluency}. For \emph{faithfulness}, each sentence is scored $0$ (refuted) or $1$ (entailed). For \emph{fluency}, scores range from $1$ (worst) to $5$ (best). We average the scores across different human evaluators for each criterion. We do not apply more fine-grained scoring scales for \emph{faithfulness}-level evaluation, as each statement in \logicnlg consists of only a single sentence.
\begin{table*}[!t]
    \centering
    \small
    \begin{tabular}{llcccc}
        \toprule
        Type & Models & SP-Acc & NLI-Acc & TAPAS-Acc & TAPEX-Acc \\
        \midrule
        \multirow{4}{*}{Fine-tuned} & 
        GPT2-C2F & 43.6 & 71.4 & 46.2 & 43.8 \\
        & R2D2 & 53.2 & 86.2 & 60.2 & 61.0\\
        & PLOG & 52.8 & 84.2 & 63.8 & \bf{69.6}\\
        & \loft & \bf{53.8} & \bf{86.6} & \bf{67.4} & 61.4\\
        \midrule
        \multirow{2}{*}{0-shot*} 
        & GPT-3.5 & \bf{54.2} & 87.6 & 81.6 & 79.4\\
        & GPT-4 & 43.2 & \bf{90.4} & \bf{91.8} & \bf{91.0}\\
        \midrule
        
        \multirow{2}{*}{1-shot Direct} &
        GPT-3.5 & \bf{60.2} & 79.0 & 80.4 & 79.2\\
        & GPT-4 & 57.6 & \bf{82.0} & \bf{87.6} & \bf{88.0}\\
        \noalign{\vskip 1ex}\cdashline{1-6}\noalign{\vskip 1ex}

        \multirow{2}{*}{1-shot CoT} &
        GPT-3.5 & 51.6 & 70.0 & 81.8 & 78.2\\
        & GPT-4 & \bf{59.8} & \bf{80.8} & \bf{89.4} & \bf{90.8}\\
        \midrule

        \multirow{11}{*}{2-shot Direct} 
        & Pythia\texttt{-12b} & 39.4 & 53.2 & 39.4 & 40.4\\
        & LLaMA\texttt{-13b} & 47.2 & 58.4 & 47.0 & 43.2\\
        & LLaMA\texttt{-7b} & 38.6 & 63.4 & 45.8 & 43.6\\
        & LLaMA2\texttt{-70b-chat} & 56.0 & 52.4 & 54.6 & 52.4\\
        & LLaMA\texttt{-30b} & 45.4 & 55.8 & 53.8 & 53.0\\
        & Alpaca\texttt{-13b} & 44.0 & 70.6 & 58.0 & 54.6\\
        & LLaMA\texttt{-65b} & 52.2 & 57.2 & 58.4 & 56.8\\
        & \tulu\texttt{-13b} & 44.4 & 68.4 & 63.4 & 59.6\\

        & Vicuna\texttt{-13b} & 51.8 & 71.4 & 66.2 & 65.2\\
        & GPT-3.5 & \bf{64.0} & 78.4 & 78.8 & 81.2\\
        & GPT-4 & 55.4 & \bf{85.8} & \bf{92.0} & \bf{89.6}\\
        \noalign{\vskip 1ex}\cdashline{1-6}\noalign{\vskip 1ex}

        \multirow{11}{*}{2-shot CoT} 
        & Pythia\texttt{-12b} & 41.8 & 54.0 & 41.2 & 42.8\\
        & LLaMA\texttt{-7b} & 38.0 & 63.2 & 48.0 & 43.0\\
        & LLaMA\texttt{-13b} & 44.2 & 53.2 & 49.2 & 48.6\\
        & LLaMA\texttt{-30b}& 45.0 & 56.6 & 60.8 & 54.2\\
        & LLaMA\texttt{-65b} & 48.0 & 58.8 & 57.4 & 57.4\\
        & \tulu\texttt{-13b} & 46.0 & 69.8 & 61.6 & 58.8\\
        & Vicuna\texttt{-13b} & 44.6 & 70.8 & 63.0 & 61.6\\
        & Alpaca\texttt{-13b} & 45.4 & 68.2 & 64.0 & 64.0\\

        & LLaMA2\texttt{-70b-chat} & 52.6 & 66.8 & 69.4 & 69.2\\
        & GPT-3.5 & 60.4 & 70.2 & 84.0 & 83.4\\
        & GPT-4 & \bf{62.2} & \bf{76.8} & \bf{88.8} & \bf{90.4}\\
        \bottomrule
    \end{tabular}
    
    \caption{Faithfulness-level automated evaluation results on the \logicnlg dataset. Within each experimental setting, we used TAPEX-Acc as the ranking indicator of model performance. $^*$: It is challenging for other LLMs to follow the instructions in 0-shot prompt to generate five statements for the input table.}
    \label{tab:logicnlg_auto_eval}
\end{table*}

\section{Experiments}
In the following subsections, we discuss the three key research questions about adopting LLMs into real-world table information seeking scenarios. Specifically, we explore LLMs' capabilities for table-to-text generation tasks, their ability to assess factual consistency, and whether they can benefit smaller fine-tuned models. The examined systems for each experiment are discussed in Appendix~\ref{sec:system}.

\subsection{RQ1: \rqone}
We experiment with two in-context learning methods, \emph{Direct Prediction} (Figure~\ref{fig:t2t_direct_prompt} in Appendix) and \emph{Chain of Thoughts} (CoT, Figure~\ref{fig:t2t_chain_prompt} in Appendix), to solve the table-to-text generation tasks. 

\paragraph{Data Insight Generation Results}
The results on the \logicnlg dataset, as displayed in Table~\ref{tab:logicnlg_auto_eval} and Table~\ref{tab:logicnlg_human_eval}, indicate that GPT-* models generally surpass the current top-performing fine-tuned models (i.e., \loft and PLOG) even in a 0-shot setting. Meanwhile, LLaMA-based models (e.g., LLaMA, Alpaca, Vicuna, \tulu) manage to achieve comparable performance to these top-performing fine-tuned models in a 2-shot setting. However, when it comes to the more challenging \lotnlg dataset, the automated evaluation result shows that only GPT-4 is capable of generating faithful statements that adhere to the specified logical reasoning types (Table~\ref{tab:lotnlg_auto_eval} in Appendix).
Moreover, increasing the number of shots or applying chain-of-thought approach does not always yield a performance gain, motivating us to explore more advanced prompting methods for data insight generation in future work.

\paragraph{Query-based Generation Results}
Table~\ref{tab:fetaqa_auto_eval} and \ref{tab:f2wtq_auto_eval} in Appendix display the automated evaluation results for the FeTaQA and \wtqnew datasets, respectively. On FeTaQA, both LLaMA-based LLM and GPT-* models achieve comparable performance to the current top-performing fine-tuned models in a 2-shot setting, indicating the capability of LLMs to answer questions requiring surface-level facts from the table. However, a significant performance gap exists between other LLMs and GPT-* models on the more challenging \wtqnew dataset. Moreover, increasing the number of shots or applying the chain-of-thought approach can both yield performance gains for query-based generation.

\begin{table}[!t]
    \centering
    \small
    \setlength\tabcolsep{3pt}
    \begin{tabular}{lcc}
        \toprule
        Model & Fluency (1-5) & Faithfulness (0-1)\\
        \midrule
        GPT2-C2F & 3.85 & 0.54 \\
        R2D2 & 4.29 & 0.72\\
        PLOG & 4.23 & 0.77\\
        \loft & 4.42 & 0.81\\
        \noalign{\vskip 1ex}\cdashline{1-3}\noalign{\vskip 1ex}
        GPT-4 0-shot & \textbf{4.82} & 0.90\\
        Vicuna 2-shot Direct & 4.69 & 0.71 \\
        Vicuna 2-shot CoT & 4.65 & 0.73 \\
        LLaMA2 2-shot Direct & 4.75 & 0.79\\
        LLaMA2 2-shot CoT & 4.70 & 0.83\\
        GPT-4 2-shot Direct & 4.71 & 0.89\\
        GPT-4 2-shot CoT & 4.77 & \textbf{0.92}\\
        \bottomrule
    \end{tabular}
    
    \caption{Human evaluation results on \logicnlg.}
    \label{tab:logicnlg_human_eval}
\end{table}
\subsection{RQ2: \rqtwo}
In RQ1, we demonstrate that LLMs can generate statements with comparative or even greater factual consistency than fine-tuned models. One natural follow-up question is whether we can employ LLMs to evaluate the faithfulness of table-to-text generation systems. This capability is crucial, as it ensures that tabular data is accurately interpreted for users, thereby preserving the credibility and reliability of real-world applications.

As discussed in Section~\ref{sec:auto_eval_definition}, existing faithfulness-level NLI-based metrics are trained on the TabFact dataset~\cite{2019TabFactA}. Recent work~\cite{chen-2023-large} has revealed that large language models using chain-of-thought prompting can achieve competitive results on TabFact. Motivated by this finding, we use the same \emph{2-shot chain-of-thought prompt} (Figure~\ref{fig:gpt_eval_prompt} in Appendix) as \citet{chen-2023-large} to generate factual consistency scores (0 for refuted and 1 for entailed) for output sentences from LogicNLG. We use GPT-3.5 and GPT-4 as the backbones, as they outperforms other LLMs in RQ1 experiments. We refer to these new metrics as \emph{CoT-3.5-Acc} and \emph{CoT-4-Acc}, respectively.

\paragraph{CoT-Acc Metrics Achieve Better Correlation with Human Judgement}\label{sec:correlation}
We leverage the human evaluation results of models (excluding GPT-4 models) in RQ1 as the \emph{human judgement}. We then compare the system-level Pearson's correlation between each evaluation metric and this human judgement.
As shown in Table~\ref{tab:human_eval_correlation}, the proposed CoT-4-Acc and CoT-3.5-Acc metrics achieve the highest and third highest correlation with human judgement, respectively. This result demonstrates LLMs' capabilities in assessing the faithfulness of table-to-text generation.
It's worth noting that although TAPAS-Acc and TAPEX-Acc perform better than CoT-4-Acc on the TabFact dataset, they exhibit lower correlation with human judgement on table-to-text evaluation. We suspect that this can be largely attributed to over-fitting on the TabFact dataset, where negative examples are created by rewriting from the positive examples. We believe that future work can explore the development of a more robust faithfulness-level metric with better alignment to human evaluation.
\begin{table}[!t]
    \centering
    \small
    \begin{tabular}{lcc}
        \toprule
        Metric & Acc on Tabfact & Pearson's correlation\\
        \midrule
        SP-Acc & 63.5 & .458\\
        NLI-Acc & 65.1 & .526\\
        TAPAS-Acc & 81.0 & .705\\
        TAPEX-Acc & \textbf{84.2} & .804\\
        \textbf{CoT-3.5-Acc} & 78.0 & .787\\
        \textbf{CoT-4-Acc} & 80.9 & \textbf{.816}\\
        \bottomrule
    \end{tabular}
    
    \caption{System-level Pearson's correlation bettwen each automated evaluation metric and human judgement. We also report the accuracy of automated evaluation metrics on the TabFact dataset for reference.}
    \label{tab:human_eval_correlation}
\end{table}
\subsection{RQ3: \rqthree}
In RQ1 and RQ2, we demonstrate the strong capability of state-of-the-art LLMs in table-to-text generation and evaluation. We next explore how fine-tuned smaller models can benefit from these abilities. We believe such exploration can provide insights for future work regarding the distillation of text generation capabilities from LLMs to smaller models~\cite{Gao2023ContinuallyIE, Scheurer2023TrainingLM, madaan2023self}. This is essential as deploying smaller, yet performance-comparable models in real-world applications could save computational resources and inference time.

\paragraph{Generating Feedback for Improving Factual Consistency}
Utilizing human feedback to enhance neural models has emerged as a significant area of interest in contemporary research~\cite{Liu2022OnIS, Gao2023ContinuallyIE, Scheurer2023TrainingLM, madaan2023self}. 
For example, \citet{Liu2022OnIS} illustrates that human-written feedback can be leveraged to improve factual consistency of text summarization systems. \citet{madaan2023self} demonstrates that LLMs can improve their initial outputs through iterative feedback and refinement. 
This work investigates whether LLMs can provide human-like feedback for outputs from fine-tuned models. Following \citet{Liu2022OnIS}, we consider generating feedback with three components: 1) \emph{Explanation}, which determine whether the initial statement is factually consistent with the given table; 2) \emph{Corrective Instruction}, which provide instructions on how to correct the initial statement if it is detected as unfaithful; and 3) \emph{Edited Statement}, which edits the initial statement following the corrective instruction. Figure~\ref{fig:feedback_prompt} in Appendix shows an example of \emph{2-shot chain-of-thought} prompts we use for feedback generation. 

\begin{table}[!t]
    \centering
    \resizebox{\linewidth}{!}{%
    \begin{tabular}{lllll}
        \toprule
        Models & TAPAS-Acc & TAPEX-Acc \\
        \midrule
        GPT2-C2F & 46.2 & 43.8\\
        \quad Edit by LLaMA2\texttt{-70b-chat} & 58.0 \up{(+11.8)} & 50.0 \up{(+6.2)}\\
        \quad Edit by GPT-3.5 & 71.0 \up{(+24.8)} & 68.4 \up{(+24.6)}\\
        \quad Edit by GPT-4 & 81.0 \up{(+34.8)} & 82.0 \up{(+38.2)} \\
        \noalign{\vskip 1ex}\cdashline{1-5}\noalign{\vskip 1ex}

        R2D2 & 60.2 & 61.0\\
        \quad Edit by LLaMA2\texttt{-70b-chat} & 65.0 \up{(+4.8)} & 60.0 \down{(-1.0)} \\
        \quad Edit by GPT-3.5 & 74.0 \up{(+13.8)} & 74.0 \up{(+13.0)}\\
        \quad Edit by GPT-4 & 87.0 \up{(+26.8)}& 89.0 \up{(+28.0)}\\
        \noalign{\vskip 1ex}\cdashline{1-5}\noalign{\vskip 1ex}
        
        PLOG & 63.8 & 69.6\\
        \quad Edit by LLaMA2\texttt{-70b-chat} & 75.0 \up{(+11.2)}& 66.0 \down{(-3.6)} \\
        \quad Edit by GPT-3.5 & 70.6 \up{(+6.8)} & 67.0 \up{(-2.6)}\\
        \quad Edit by GPT-4 & 91.0 \up{(+27.2)} & 86.0 \up{(+16.4)} \\
        \noalign{\vskip 1ex}\cdashline{1-5}\noalign{\vskip 1ex}
        
        \loft & 67.4 & 61.4\\
        \quad Edit by LLaMA2\texttt{-70b-chat} & 72.0 \up{(+4.6)} & 64.0 \up{(+2.6)} \\
        \quad Edit by GPT-3.5 & 70.0 \up{(+2.6)} & 65.6 \up{(+4.2)}\\
        \quad Edit by GPT-4 & 81.0 \up{(+13.6)} & 86.0 \up{(+24.6)}\\
        \bottomrule
    \end{tabular}
    }
    \caption{Automated evaluation results on \logicnlg using statements pre-edited and post-edited by LLMs.}
    \label{tab:feedback_auto_eval}
\end{table}
\paragraph{Feedback from LLMs is of High Quality}
We assess the quality of generated feedback through automated evaluations.
Specifically, we examine the faithfulness scores of \emph{Edited Statements} in the generated feedback, comparing these scores to those of the original statements. 
We report TAPAS-Acc and TAPEX-Acc for experimental results, as these two metrics exhibit better alignment with human evaluation (Section~\ref{sec:correlation}). As illustrated in Table~\ref{tab:feedback_auto_eval}, LLMs can effectively edit statements to improve their faithfulness, particularly for outputs from lower-performance models, such as GPT2-C2F. 
\section{Related Work}
\paragraph{Table-to-Text Generation}
Text generation from semi-structured knowledge sources, such as web tables, has been studied extensively in recent years~\cite{parikh-etal-2020-totto, chen-etal-2020-logical, cheng-etal-2022-hitab, zhao-etal-2023-openrt}. The goal of the table-to-text generation task is to generate natural language statements that faithfully describe information contained in the provided table region. 
The most popular approach for table-to-text generation tasks is to fine-tune a pre-trained language model on a task-specific dataset~\cite{chen-etal-2020-logical,liu-etal-2022-plog, zhao-etal-2022-reastap, nan-etal-2022-r2d2, zhao-etal-2023-loft}. 
To the best of our knowledge, we are the first to systematically evaluate the performance of LLMs on table-to-text generation tasks.
% \citet{zhao2023qtsumm} also evaluates the GPT-3 zero- and few-shot performance on the query-focused table summarization task.

\paragraph{Large Language Models}
LLMs have demonstrated remarkable in-context learning capabilities~\cite{gpt-3, Chowdhery2022PaLMSL, scao2022bloom,Chung2022ScalingIL, OpenAI2023GPT4TR}, where the model receives a task demonstration in natural language accompanied by a limited number of examples. The Chain-of-Thought prompting methods~\cite{wei2022chain, Wang2022SelfConsistencyIC} further empower LLMs to perform complex reasoning tasks~\cite{Han2022FOLIONL, zhao2023qtsumm, Ye2023LargeLM, chen-2023-large}. 
More recent works~\cite{chen-2023-large, nan2023enhancing} investigate in-context learning capabilities of LLMs on table-based tasks, including table question answering~\cite{pasupat-liang-2015-compositional, iyyer-etal-2017-search, zhong2018seqsql} and table fact checking~\cite{2019TabFactA, gupta-etal-2020-infotabs}. 
However, the potential of LLMs in generating text from tabular data remains underexplored. 

\section{Conclusion}
This paper investigates the potential of applying LLMs in real-world table information seeking scenarios. We demonstrate their superiority in faithfulness, and their potential as evaluation systems. Further, we provide valuable insights into leveraging LLMs to generate high-fidelity natural language feedback. We believe that the findings of this study could benefit real-world applications, aimed at improving user efficiency in data analysis.

\section*{Ethical Consideration}
\lotnlg and \wtqnew were constructed upon the test set of \logicnlg~\cite{chen-etal-2020-logical} and WTQ~\cite{pasupat-liang-2015-compositional} datasets, which are publicly available under the licenses of MIT\footnote{\url{https://opensource.org/licenses/MIT}} and CC BY-SA 4.0\footnote{\url{https://creativecommons.org/licenses/by-sa/4.0/}}, respectively. These licenses permit us to modify, publish, and distribute additional annotations upon the original dataset. 

\bibliography{anthology,custom}
\bibliographystyle{acl_natbib}

\appendix
\section{Table-to-Text Generation Benchmarks} \label{appendix:dataset}
\subsection{\lotnlg Dataset}\label{appendix:type}
\paragraph{Logical Reasoning Type Definition}
\begin{itemize} [leftmargin=*]
\itemsep0em  
\item Aggregation: operations involving sum or average operation to summarize the overall statistics. Sentence: The total number of scores of xxx is xxx. The average value of xxx is xxx.
\item Negation: operations to negate. Sentence: xxx did not get the first prize.
\item Superlative: superlative operations to get the highest or lowest value. Sentence: xxx achieved the most scores.
\item Count: operations to count the amount of entities that fulfil certain conditions. Sentence: There are 4 people born in xxx.
\item Comparative: operations to compare a specific aspect of two or more entities. Sentence: xxx is taller than xxx.
\item Ordinal: operations to identify the ranking of entities in a specific aspect. Sentence: xxx is the third youngest player in the game.
\item Unique: operations to identify different entities. Sentence: The players come from 7 different cities.
\item All: operations to summarize what all entities do/have in common. Sentence: All of the xxx are more expensive than \textdollar25.
\item Surface-Level: no logical reasoning type above. Sentence: xxx is moving to xxx.
\end{itemize}

\begin{figure}[h]
    \centering
    \includegraphics[width = 0.9\linewidth]{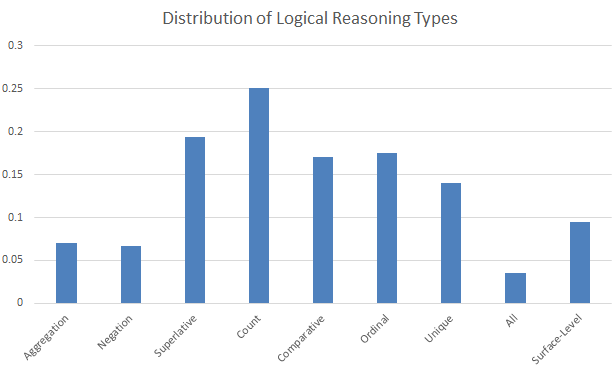}
    \caption{Distribution of logical reasoning types for the \lotnlg dataset.}
    \label{fig:type_distribution}
\end{figure}

\section{Examined Systems}\label{sec:system}
\subsection{Fine-tuned Models}
\begin{itemize}[leftmargin=*]
    \itemsep0em 
    \item \textbf{BART}~\cite{lewis-etal-2020-bart} is a pre-trained denoising autoencoder with transformer-based architecture and shows effectiveness in NLG tasks.
    % \item \textbf{T5}~\cite{2020t5} demonstrates effectiveness in NLG tasks by treating all NL problems as text-to-text tasks during pre-training stage.
    \item \textbf{Flan-T5}~\cite{Chung2022ScalingIL} enhances T5~\cite{2020t5} by scaling instruction fine-tuning and demonstrates better human-like reasoning abilities than the T5.

\item\textbf{GPT2-C2F}~\cite{chen-etal-2020-logical} first generates a template which determines the global logical structure, and then produces the statement using the template as control.

\item\textbf{R2D2}~\cite{nan-etal-2022-r2d2} trains a generative language model both as a generator and a faithfulness discriminator with additional replacement detection and unlikelihood learning tasks, to enhance the faithfulness of table-to-text generation.

\item\textbf{TAPEX}~\cite{liu2022tapex} continues pre-training the BART model by using a large-scale corpus of synthetic SQL query execution data, showing better table understanding and reasoning abilities.

\item \textbf{OmniTab}~\cite{jiang-etal-2022-omnitab} uses the same backbone as TAPEX, and is further pre-trained on collected natural and synthetic Table QA examples. 

\item\textbf{ReasTAP}~\cite{zhao-etal-2022-reastap} enhances the table understanding and reasoning abilities of BART by pre-training on a synthetic Table QA corpus. 

\item\textbf{PLOG}~\cite{liu-etal-2022-plog} continues pre-training text generation models on a table-to-logic-form generation task (i.e., T5 model), improving the faithfulness of table-to-text generation. \item\textbf{\loft}~\cite{zhao-etal-2023-loft} utilizes logic forms as fact verifiers and content planners to control table-to-text generation, exhibiting improved faithfulness and text diversity. 
\end{itemize}

\subsection{Large Language Models}
\begin{itemize}[leftmargin=*]
    \item \textbf{Pythia}~\cite{biderman2023pythia} is a suite of 16 open-sourced LLMs all trained on public data in the exact same order and ranging in size from 70M to 12B parameters. This helps researchers to gain a better understanding of LLMs and their training dynamics.
    \item \textbf{LLaMA}~\cite{touvron2023llama, Touvron2023Llama2O} is an open-source LLM trained on large-scale and publicly available datasets. We evaluate both LLaMA and LLaMA2 in this paper.
    \item \textbf{Alpaca}~\cite{alpaca} and \textbf{Vicuna}~\cite{vicuna2023} are fine-tuned from LLaMA with instruction-following data, exhibiting better instruction-following capabilities. 
    \item \textbf{\tulu}~\cite{wang2023tulu} further trains LLaMA on 12 open-source instruction datasets, achieving better performance than LLaMA.
    \item \textbf{GPT}~\cite{gpt-3, wei2022chain} is a powerful large language model which is capable of generating human-like text and performing a wide range of NLP tasks in a few-shot setting. We use the OpenAI engines of \texttt{gpt-3.5-0301} and \texttt{gpt-4-0314} for GPT-3.5 and GPT-4 models, respectively.
\end{itemize}

To formulate the prompt, we linearize the table as done in previous work on table reasoning~\cite{chen-2023-large} and concatenate it with its corresponding reference statements as demonstrations. We use the table truncation strategy as proposed by \citet{liu2022tapex} to truncate large table and ensure that the prompts are within the maximum token limitation for each type of LLMs. For LLM parameter settings, we used a temperature of 0.7, maximum output length of 512, without any frequency or presence penalty. 
\section{Experiments}
\begin{figure}[h]
    \centering
    \includegraphics[width = \linewidth]{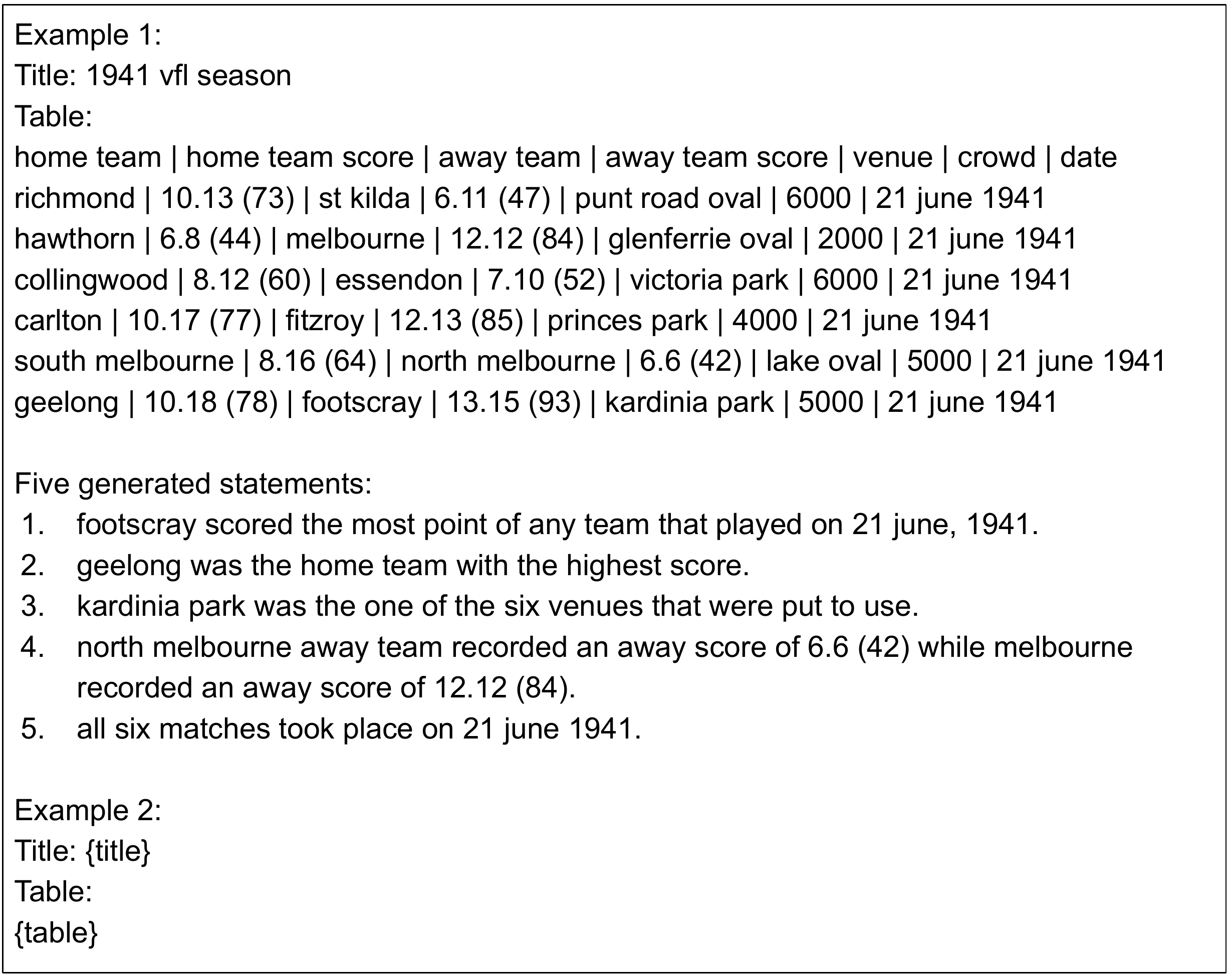}
    \caption{An example of 1-shot \emph{direct-prediction} prompting for the \logicnlg task.}
    \label{fig:t2t_direct_prompt}
\end{figure}

\begin{figure}[h]
    \centering
    \includegraphics[width = \linewidth]{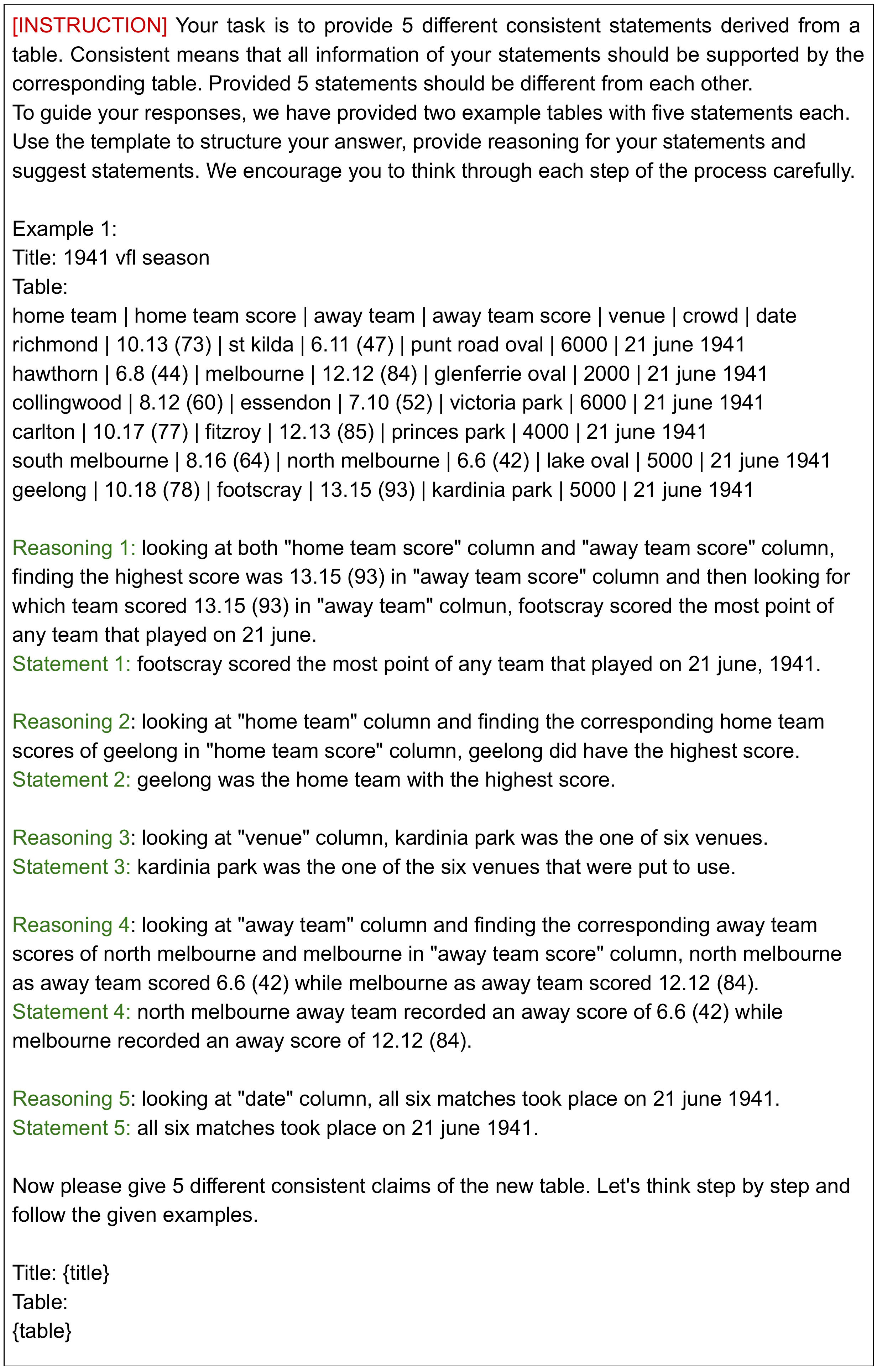}
    \caption{An example of 1-shot \emph{chain-of-thought} prompting for the \logicnlg task.}
    \label{fig:t2t_chain_prompt}
\end{figure}

\begin{table*}[!t]
    \centering
    \small
    \begin{tabular}{llcccccc}
        \toprule
        Type & Models & SP-Acc & NLI-Acc & TAPAS-Acc & TAPEX-Acc & Type EM & Type F1\\
        \midrule
        \multirow{2}{*}{0-shot*} 
        & GPT-3.5 & 51.2 & 77.2 & 70.8 & 66.8 & 59.2 & 43.8\\
        & GPT-4 & \bf{69.2} & \bf{79.4} & \bf{85.6} & \bf{84.2} & \bf{75.2} & \bf{60.0}\\
        \midrule
        
        \multirow{2}{*}{1-shot Direct}
        & GPT-3.5 & 53.8 & 75.6 & 71.6 & 71.0 & 51.2 & 38.1\\
        & GPT-4 & \bf{60.2} & \bf{72.8} & \bf{83.8} & \bf{84.2} & \textbf{76.6} & \textbf{63.0}\\
        \noalign{\vskip 1ex}\cdashline{1-8}\noalign{\vskip 1ex}

        \multirow{2}{*}{1-shot CoT}
        & GPT-3.5 & 50.8 & \bf{78.8} & 79.2 & 79.4 & 46.2 & 30.2\\
        & GPT-4 & \bf{59.2} & 74.8 & \bf{84.4} & \bf{85.8} & \bf{70.0} & \bf{51.6} \\
        \midrule

        \multirow{11}{*}{2-shot Direct} 
        & Pythia\texttt{-12b} & 44.2 & 60.6 & 41.8 & 43.0 & 19.0 & 12.2\\
        & LLaMA\texttt{-7b} & 41.0 & 62.2 & 46.2 & 46.2 & 18.2 & 13.4\\
        & Vicuna\texttt{-13b} & 48.6 & 71.2 & 57.4 & 54.4 & 22.0 & 15.2\\
        & LLaMA\texttt{-13b} & 44.6 & 62.4 & 50.8 & 48.8 & 22.6 & 15.8\\
        & Alpaca\texttt{-13b} & 46.2 & 73.8 & 50.8 & 54.0 & 21.8 & 15.8\\
        & LLaMA2\texttt{-70b-chat} & 44.2 & 60.0 & 56.0 & 58.0 & 24.2 & 15.8\\
        & LLaMA\texttt{-30b} & 40.0 & 62.6 & 53.0 & 52.6 & 24.2 & 16.4\\
        & LLaMA\texttt{-65b} & 46.2 & 57.8 & 54.0 & 51.8 & 21.0 & 17.2\\
        & \tulu\texttt{-13b} & 44.2 & 72.8 & 60.8 & 56.8 & 26.6 & 17.4\\

        & GPT-3.5 & 55.2 & \bf{76.2} & 70.8 & 67.6 & 52.2 & 35.0\\
        & GPT-4 & \bf{61.4} & 72.2 & \bf{84.6} & \bf{83.2} & \bf{73.4} &\bf{54.8}\\
        \noalign{\vskip 1ex}\cdashline{1-8}\noalign{\vskip 1ex}

        \multirow{11}{*}{2-shot CoT} 
        & Pythia\texttt{-12b} & 42.0 & 53.8 & 41.2 & 41.0 & 15.2 & 11.6\\
        & LLaMA\texttt{-30b}& 41.0 & 60.4 & 52.6 & 59.2 & 20.4 & 13.2\\
        & LLaMA\texttt{-7b} & 37.6 & 61.2 & 43.8 & 45.0 & 17.2 & 13.4\\
        & LLaMA2\texttt{-70b-chat} & 48.2 & 64.6 & 56.0 & 67.8 & 20.2 & 13.4\\
        & LLaMA\texttt{-13b} & 45.0 & 56.6 & 51.2 & 51.2 & 18.8 & 14.0\\
        & LLaMA\texttt{-65b} & 45.2 & 62.4 & 59.4 & 58.8 & 21.2 & 15.2\\
        & Vicuna\texttt{-13b} & 43.4 & 72.0 & 62.2 & 61.0 & 18.4 & 16.0\\
        & Alpaca\texttt{-13b} & 40.4 & 71.6 & 58.4 & 57.8 & 23.0 & 16.2\\
        & \tulu\texttt{-13b} & 45.8 & 65.8 & 60.8 & 61.0 & 23.2 & 16.2\\
    
        & GPT-3.5 & 49.2 & \bf{74.4} & 77.2 & 75.4 & 49.4 & 35.0\\
        & GPT-4 & \bf{59.2}& 72.0 & \bf{85.6} & \bf{83.2} & \bf{67.6} & \bf{55.6}\\
        \bottomrule
    \end{tabular}
    \caption{Faithfulness-level automated evaluation results on \lotnlg. We do not evaluate fine-tuned models as \lotnlg does not contain a training set. $^*$: It is challenging for other LLMs to follow the instructions in 0-shot prompt to generate a statement using the specified types of logical reasoning operations.}
    \label{tab:lotnlg_auto_eval}
\end{table*}
\begin{table*}[!t]
    \centering
    \small
    \begin{tabular}{llcccc}
        \toprule
        Type & Models & BLEU-1/2/3 & ROUGE-1/2/L & TAPAS-Acc & TAPEX-Acc\\
        \midrule
        \multirow{5}{*}{Fine-tuned} & 
        BART & 63.2/50.8/42.0 & \textbf{67.6}/\textbf{46.0}/\textbf{57.2} & 94.8 & 68.8\\
        & Flan-T5 & 62.2/49.6/41.0 & 66.8/45.0/56.2 & 94.2 & 69.2\\
        & OmniTab & 63.4/50.8/41.8 & 67.4/45.2/56.2 & 94.6 & 71.6\\
        & ReasTAP & \textbf{63.6}/\textbf{51.0}/\textbf{42.2} & \textbf{67.6}/45.8/\textbf{57.2} & 94.6 & 71.4\\
        & TAPEX & \textbf{63.6}/50.8/42.0 & 66.4/45.0/56.2 & \textbf{96.2} & \textbf{73.0}\\
        \midrule
        \multirow{2}{*}{0-shot}
        & GPT-3.5 & \textbf{56.4}/\textbf{42.6}/\textbf{33.4} & 60.6/38.0/49.4 & 92.4 & 72.8\\
        & GPT-4 & 52.4/40.2/31.8 & \textbf{63.8}/\textbf{40.4}/\textbf{51.6} & \textbf{94.0} & \textbf{74.4}\\
        \midrule
        
        \multirow{2}{*}{1-shot Direct}
        & GPT-3.5 & \textbf{56.8}/43.2/34.2 & 63.0/39.8/51.4 & 91.8 & \textbf{74.6}\\
        & GPT-4 & 56.4/\textbf{43.6}/\textbf{34.8} & \textbf{66.2}/\textbf{43.0}/\textbf{54.4} & \textbf{94.0} & 73.8\\
        \noalign{\vskip 1ex}\cdashline{1-6}\noalign{\vskip 1ex}

        \multirow{2}{*}{1-shot CoT}
        & GPT-3.5 & 43.2/32.4/25.2 & 57.4/35.8/46.8 & \textbf{94.2} & 67.0\\
        & GPT-4 & \textbf{59.6}/\textbf{45.8}/\textbf{36.4} & \textbf{64.0}/\textbf{41.0}/\textbf{52.4} & 91.0 & \textbf{76.4}\\
        \midrule

        \multirow{11}{*}{2-shot Direct} 
        & Pythia\texttt{-12b} & 38.8/26.6/19.4 & 43.2/22.6/35.2 & 76.6 & 35.0\\
        & LLaMA\texttt{-7b} & 40.6/28.6/21.4 & 48.2/26.6/39.0 & 86.2 & 47.8\\
        & LLaMA\texttt{-13b} & 48.4/35.2/26.8 & 51.0/29.4/42.2 & 85.4 & 57.4\\
        & Alpaca\texttt{-13b} & 52.2/38.4/29.6 & 56.4/33.6/46.2 & 88.4 & 57.4\\
        
        & \tulu\texttt{-13b} & 50.6/37.4/29.0 & 54.2/31.8/44.6 & 86.4 & 60.0\\
        & LLaMA\texttt{-30b} & 50.4/37.0/28.2 & 56.2/33.2/45.4 & 87.0 & 60.2\\
        & Vicuna\texttt{-13b} & 56.0/42.2/32.8 & 59.0/36.2/48.0 & 87.6 & 62.4\\
        & LLaMA\texttt{-65b} & 53.6/39.8/30.8 & 57.0/34.0/46.6 & 88.4 & 63.0\\
        & LLaMA2\texttt{-70b-chat} & 54.6/41.0/31.8 & 58.4/35.8/47.8 & 89.4 & 66.2\\
        
        & GPT-4 & 55.0/\textbf{42.8}/\textbf{34.6} & \textbf{66.0}/\textbf{42.8}/\textbf{54.0} & \textbf{95.2} & 75.8\\
        & GPT-3.5 & \textbf{55.8}/\textbf{42.8}/34.0 & 63.2/40.0/51.6 & 92.2 & \textbf{76.0}\\
        \noalign{\vskip 1ex}\cdashline{1-6}\noalign{\vskip 1ex}

        \multirow{11}{*}{2-shot CoT} 
        & Pythia\texttt{-12b} & 38.8/25.4/17.8 & 39.2/18.8/32.2 & 69.0 & 36.2\\
        & LLaMA\texttt{-7b} & 33.0/22.2/16.0 & 41.0/21.2/33.2 & 77.6 & 42.0\\
        & LLaMA\texttt{-13b} & 43.2/30.4/22.6 & 45.4/25.2/37.6 & 82.0 & 50.8\\
        & Alpaca\texttt{-13b} & 47.4/34.4/26.2 & 51.4/30.0/42.0 & 82.8 & 54.4\\
        & \tulu\texttt{-13b} & 37.0/25.8/18.8 & 43.6/24.0/35.2 & 86.2 & 55.8\\

        & LLaMA\texttt{-30b} & 45.4/33.2/25.6 & 52.4/30.8/42.2 & 86.2 & 63.6\\
        & Vicuna\texttt{-13b} & 50.4/37.6/29.4 & 53.8/32.4/44.6 & 85.6 & 65.8\\
        & LLaMA\texttt{-65b} & 50.2/37.0/28.4 & 54.8/32.8/44.6 & 87.8 & 66.0\\
        & LLaMA2\texttt{-70b-chat} & 53.8/40.2/31.4 & 57.4/34.8/47.0 & 89.2 & 66.2\\
        & GPT-3.5 & 50.8/38.8/30.8 & 60.6/38.2/49.0 & \textbf{92.8} & 70.8\\
        & GPT-4 & \textbf{62.2}/\textbf{48.6}/\textbf{39.2} & \textbf{65.8}/\textbf{42.8}/\textbf{54.4} & 91.2 & \textbf{79.2}\\
        
        \bottomrule
    \end{tabular}
    
    \caption{Automated evaluation results on the FeTaQA dataset.}
    \label{tab:fetaqa_auto_eval}
\end{table*}
\begin{table*}[!t]
    \centering
    \small
    \begin{tabular}{llrrccc}
        \toprule
        Type & Models & BLEU-1/2/3 & ROUGE-1/2/L & TAPAS-Acc & TAPEX-Acc & Accuracy\\
        \midrule
        \multirow{2}{*}{0-shot}
        & GPT-3.5 & \textbf{63.2}/\textbf{49.2}/\textbf{39.4} & 64.4/40.0/\textbf{56.4} & 73.0 & 74.6 & 54.0\\
        & GPT-4 & 60.6/46.8/37.4 & \textbf{64.6}/\textbf{40.4}/54.8 & \textbf{78.6} & \textbf{80.6} & \textbf{62.4}\\
        \midrule
        
        \multirow{2}{*}{1-shot Direct}
        & GPT-3.5 & 62.0/48.4/39.0 & 64.0/40.0/56.8 & 75.0 & 73.2 & 51.8\\
        & GPT-4 & \textbf{63.2}/\textbf{49.8}/\textbf{40.4} & \textbf{66.2}/\textbf{42.6}/\textbf{58.0} & \textbf{78.4} & \textbf{79.0} & \textbf{66.0}\\
        \noalign{\vskip 1ex}\cdashline{1-7}\noalign{\vskip 1ex}

        \multirow{2}{*}{1-shot CoT}
        & GPT-3.5 & 55.0/42.4/33.8 & 62.8/39.0/54.8 & 72.4 & 72.2 & 55.2\\
        & GPT-4 & 62.2/49.0/39.6 & \textbf{66.2}/\textbf{42.2}/\textbf{58.4} & \textbf{78.2} & \textbf{78.6} & \textbf{69.8}\\
        \midrule

        \multirow{11}{*}{2-shot Direct} 
        & Pythia\texttt{-12b} & 12.4/7.6/5.2 & 19.6/9.2/17.4 & 74.6 & 62.4 & 7.8 \\
        & LLaMA\texttt{-7b} & 14.4/9.6/6.8 & 26.2/13.4/23.0 & 71.8 & 53.0 & 19.0\\
        & LLaMA\texttt{-13b} & 7.6/4.8/3.4 & 20.2/10.4/18.2 & 78.4 & 56.0 & 21.4\\
        & Vicuna\texttt{-13b} & 43.0/31.6/24.4 & 46.0/27.2/40.6 & 74.6 & 64.2 & 30.2\\
        & Alpaca\texttt{-13b} & 40.8/29.2/21.6 & 46.6/26.2/40.4 & 71.8 & 57.6 & 31.2\\
        & LLaMA\texttt{-30b} & 34.0/24.4/18.2 & 44.6/25.0/39.8 & 74.0 & 61.0 & 31.8\\
        & \tulu\texttt{-13b} & 49.6/36.4/28.0 & 51.4/29.4/45.8 & 78.8 & 60.4 & 33.8\\
        
        & LLaMA\texttt{-65b} & 45.8/33.8/26.0 & 48.8/28.2/43.6 & 73.6 & 64.4 & 36.2\\
        & LLaMA2\texttt{-70b-chat} & 51.2/38.4/30.0 & 50.4/29.6/45.4 & 72.4 & 68.4 & 37.6\\
        & GPT-3.5 & \textbf{63.4}/\textbf{49.8}/40.2 & 64.8/40.8/57.2 & 74.8 & 73.6 & 51.8\\
        & GPT-4 & 62.8/49.2/39.6 & \textbf{65.8}/\textbf{41.8}/\textbf{57.6} & \textbf{78.6} & \textbf{81.4} & \textbf{63.6}\\
        \noalign{\vskip 1ex}\cdashline{1-7}\noalign{\vskip 1ex}

        \multirow{11}{*}{2-shot CoT} 
        & Pythia\texttt{-12b} & 27.2/18.0/12.8 & 35.6/17.4/31.4 & 66.0 & 48.8 & 15.8\\
        & LLaMA\texttt{-7b} & 13.2/8.4/5.8 & 28.0/13.2/24.0 & 73.4 & 47.8 & 24.2\\
        & LLaMA\texttt{-13b} & 22.2/14.8/10.4 & 35.2/18.0/31.4 & 74.0 & 56.2 & 26.2\\
        & Alpaca\texttt{-13b} & 33.2/23.6/17.8 & 47.6/26.4/41.2 & 75.0 & 55.4 & 32.2\\
        & LLaMA\texttt{-30b} & 37.4/26.2/19.6 & 46.2/24.8/40.6 & 72.6 & 60.0 & 35.6\\
        & \tulu\texttt{-13b} & 25.8/17.0/12.0 & 35.4/17.4/31.0 & \textbf{79.0} & 65.6 & 35.8\\
        & Vicuna\texttt{-13b} & 45.2/33.2/25.4 & 53.6/31.2/47.6 & 75.6 & 62.2 & 38.6\\
        & LLaMA\texttt{-65b} & 51.2/37.8/29.0 & 51.6/29.4/45.6 & 75.6 & 67.6 & 41.6\\
        & LLaMA2\texttt{-70b-chat} & 46.2/34.2/26.6 & 49.6/28.8/44.2 & 75.8 & 66.6 & 43.2\\

        & GPT-3.5 & 57.4/44.4/35.4 & 64.0/40.0/55.4 & 73.6 & 72.8 & 58.6\\
        & GPT-4 & \textbf{63.0}/\textbf{49.6}/\textbf{40.0} & \textbf{66.2}/\textbf{42.4}/\textbf{58.8} & 76.4 & \textbf{79.6} & \textbf{68.4}\\

        \bottomrule
    \end{tabular}
    \caption{Automated evaluation results on the \wtqnew dataset. We do not evaluate fine-tuned models as \wtqnew does not contain a training set.}
    \label{tab:f2wtq_auto_eval}
\end{table*}

\begin{figure}[h]
    \centering
    \includegraphics[width = \linewidth]{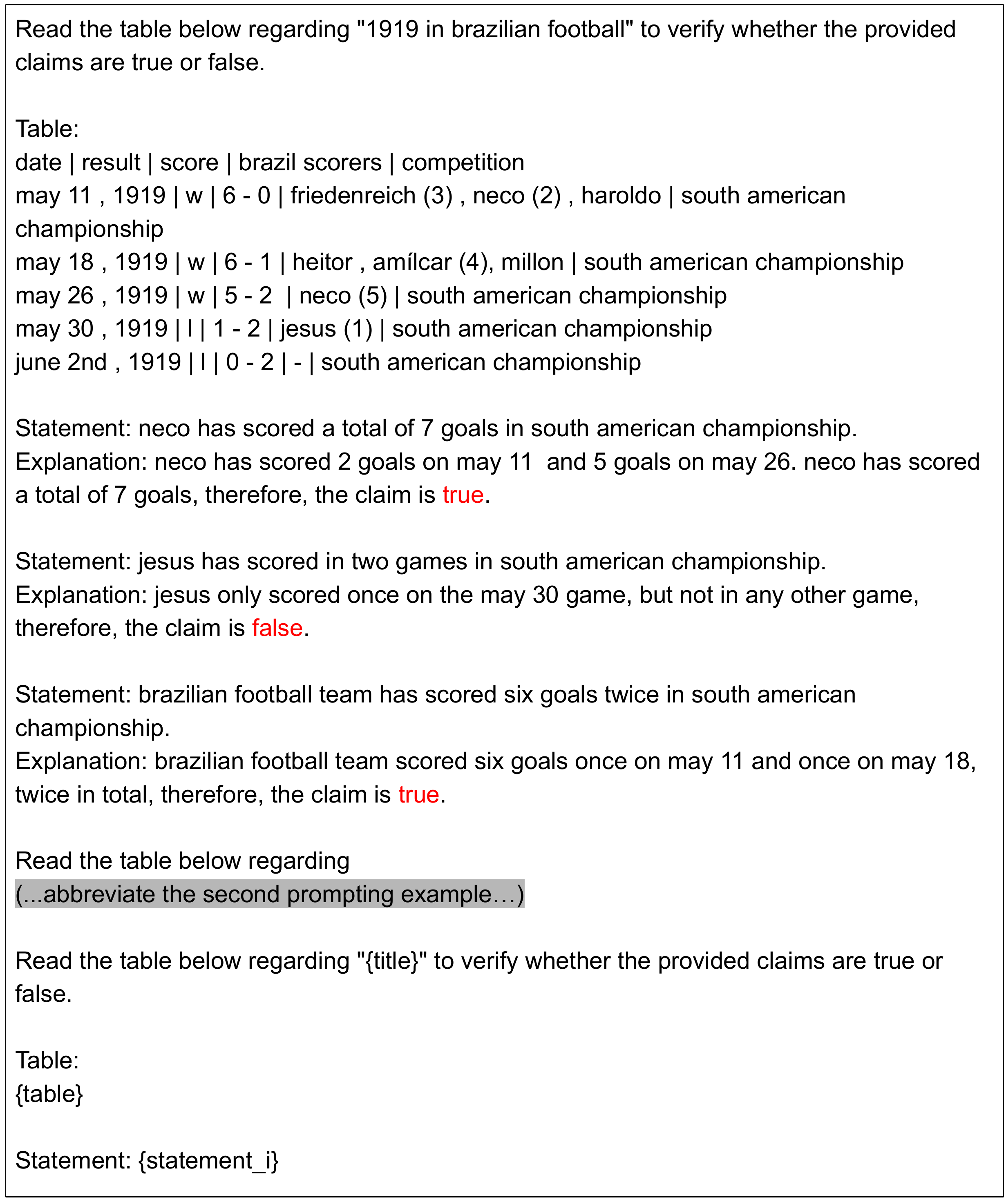}
    \caption{An example of 2-shot chain-of-thought prompting adopted from \citet{chen-2023-large} for faithfulness-level automated evaluation.}
    \label{fig:gpt_eval_prompt}
\end{figure}
\begin{figure*}[h]
    \centering
    \includegraphics[width = 0.7\linewidth]{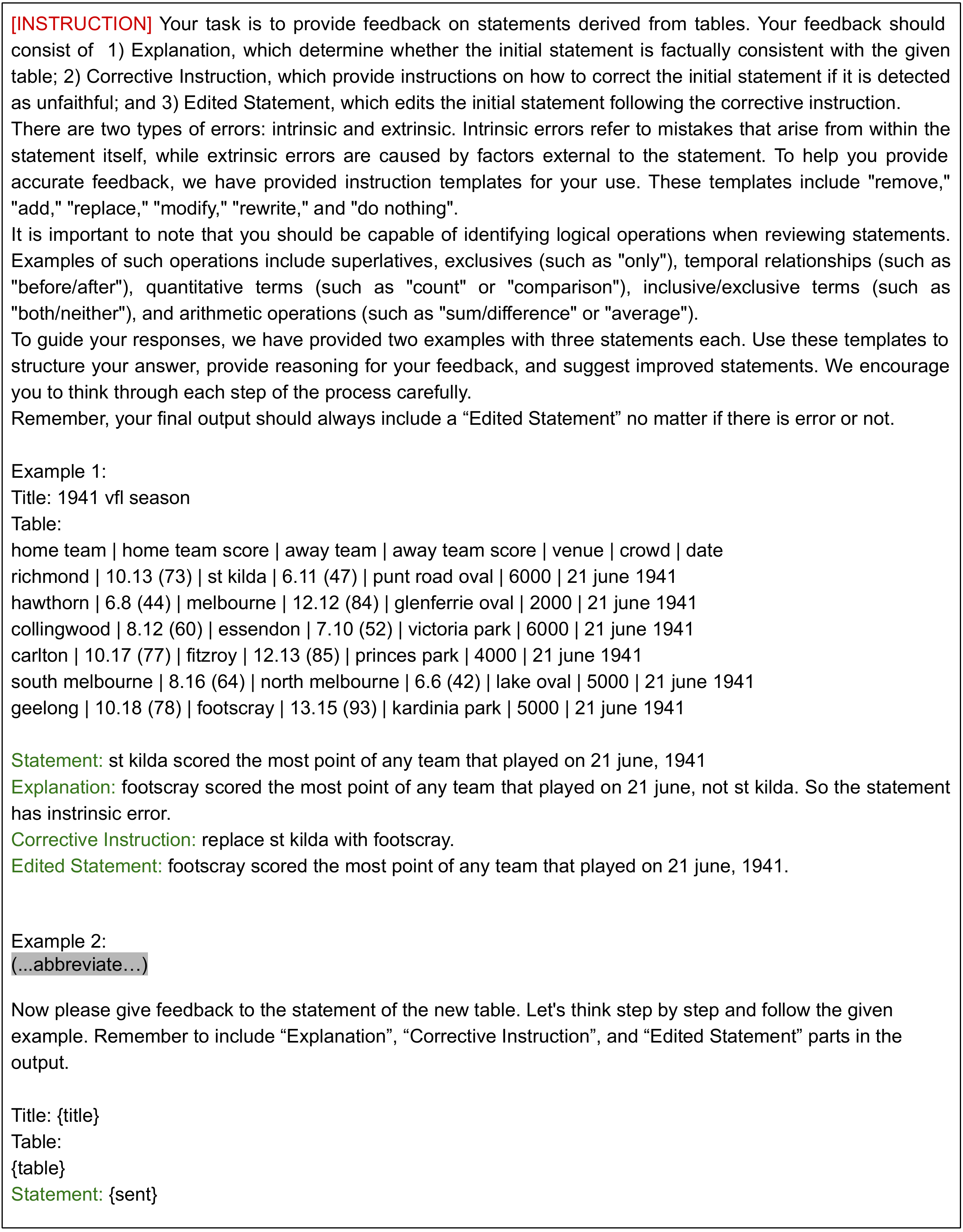}
    \caption{An example of 2-shot \emph{chain-of-thought} prompts for natural language feedback generation on \logicnlg.}
    \label{fig:feedback_prompt}
\end{figure*}

 \end{document}